\documentclass[Journal,SingleSpace,InsideFigs,NoLineNumbers,letterpaper]{dronar-preprint}
\usepackage[utf8]{inputenc}
\usepackage[T1]{fontenc}
\usepackage{lmodern}
\usepackage{graphicx}
\usepackage[section]{placeins}
\usepackage{float}
\usepackage{needspace}
\usepackage{etoolbox}
\pretocmd{\section}{\Needspace{6\baselineskip}}{}{}
\pretocmd{\subsection}{\Needspace{5\baselineskip}}{}{}
\pretocmd{\subsubsection}{\Needspace{4\baselineskip}}{}{}
\usepackage[style=base,figurename=Fig.,labelfont=bf,labelsep=period]{caption}
\usepackage{subcaption}
\usepackage{amsmath}
\usepackage{newtxtext,newtxmath}
\usepackage[colorlinks=true,citecolor=red,linkcolor=black]{hyperref}
\NameTag{Zielinski et al., September 2026}
\begin{document}

\title{Non-Invasive Inspection of Water Canals Using Dronar}

\author[1]{Michael Zielinski}
\author[1]{Zhizhan Wang}
\author[2]{Benjamin Z. Dymond, Ph.D.}
\author[1]{Reza Razavian, Ph.D.}
\author[3]{Zhongwang Dou, Ph.D.}

\affil[1]{Mechanical Engineering Department, Steve Sanghi College of Engineering, Northern Arizona University, Flagstaff, AZ 86011, USA. Email: msz32@nau.edu}
\affil[2]{Civil and Environmental Engineering Department, Steve Sanghi College of Engineering, Northern Arizona University, Flagstaff, AZ 86011, USA}
\affil[3]{Department of Biomedical Engineering, Kate Gleason College of Engineering, Rochester Institute of Technology, Rochester, NY 14623, USA}

\maketitle

\begin{abstract}
Open concrete canals play a vital role in water transportation, serving as primary water infrastructure for millions of people across the Phoenix, Arizona, metro area. Over time, the concrete canals can experience a range of issues, including canal lining deformation, cracked concrete, and sediment buildup on the canal floor. Identifying such critical issues is a resource-intensive process, which currently happens only during four-year dry-up cycles. This prevents the maintenance crew from prioritizing operations on the most affected canal segments. To address this issue, the research team has developed and verified an easily deployable and non-invasive method to inspect canal beds without draining the water. This inspection system integrates affordable, off-the-shelf drone and sonar technology (termed \textit{dronar}). This dronar system includes a consumer-grade sonar system integrated into an unmanned surface vehicle (USV) that carries the sonar transducer just under the surface of the canal water. This paper presents a proof-of-concept demonstration of the dronar system across three field tests on the Arizona Canal in Phoenix. DownScan depth profiles from the sedimented canal segment were consistently shallower than profiles from the same segment after cleaning, with offsets of up to 15 cm observed along the track. Repeated runs over the clean segment produced closely overlapping DownScan depth profiles, confirming that the dronar yields repeatable measurements across the natural variation of the canal bed. These results establish the dronar as a viable proof-of-concept tool for non-invasive canal bed inspection.
\end{abstract}

\section{Introduction and Background}

\begin{figure}[H]
\includegraphics[width=10.0cm]{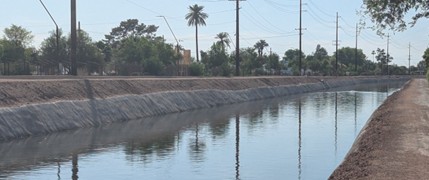}
\caption{Example concrete water canal in the SRP system.\label{fig:canal}}
\end{figure}

In the Salt River Valley of central Arizona, the Salt River Project (SRP) manages a water canal system consisting of seven canals totaling over 211 km; an example concrete canal is shown in Figure~\ref{fig:canal}. These canals are the primary water resource for nearly 2.5 million homes in Phoenix, AZ. Beyond household and industrial usage, this canal system is also the primary water source for Phoenix's agriculture. Over time, concrete canals can experience a range of issues, including canal lining deformation, cracked concrete, and sediment buildup on the canal floor. These issues reduce the efficiency of water transport and increase operating and maintenance costs \cite{ref-paudel2010:2010a}. Currently, SRP \textit{zanjeros} (``ditch riders'') can only identify deformation/cracks located above the water level using visual inspection. In addition, many cracks in the canal concrete are often covered by sediment, which makes visual identification nearly impossible without draining the canal. Thus, every fall and winter, segments of SRP's major canals are emptied (dry-up) to allow for construction, cleanup, and repairs. However, due to the high cost of the dry-up procedure and the considerable size of the canal system, each canal segment is only maintained via dry-up once every four years. Considering these challenges, the primary objective of this project was to develop a non-invasive canal inspection method using affordable, off-the-shelf drone and sonar technology (termed \textit{dronar}) to inspect canal beds without draining the water. The benefits of this method include (i) the location and characterization of the level of sediment buildup, which can prioritize the locations for maintenance crews to expend their resources and (ii) the location of potential canal liner deformations and concrete cracks before severe leakage occurs, which supports the scheduling of targeted maintenance.

Sonar mapping is the most widely adopted technique for exploring and mapping underwater environments because sound waves travel farther in water than radar and light waves \cite{ref-noaa-sonar}. Historically, sonar-based investigation has required specialized vessels and trained operators, but the performance of consumer-grade ``fish finder'' sonar systems has improved considerably in recent years \cite{ref-dainys2022}. Johnson and Helferty established the foundational principles for interpreting side-scan sonar imagery in terms of seafloor material properties, demonstrating that acoustic backscatter can be used to characterize bottom composition and texture \cite{ref-johnson1990}. Building on this foundation, Yamasaki et al. demonstrated that a modern consumer-grade fish finder mounted on a remotely operated watercraft can successfully characterize water-bottom topography and sediment conditions in shallow freshwater environments, validating the use of affordable sonar hardware for scientific investigation \cite{ref-yamasaki2017}. Furthermore, a few research groups have demonstrated that it is feasible to integrate lightweight consumer-grade sonar with unmanned aerial vehicles for shallow-water surveys. Diaz et al. and Ruffell et al. demonstrated centimeter-scale depth accuracy in shallow-water bathymetric surveys using aerial drone-tethered consumer sonar systems \cite{ref-diaz2022,ref-ruffell2021}. Greene et al. and Powers et al. further showed that side-scan sonar can identify and characterize submerged features in water depths as shallow as one meter, including seagrass beds and freshwater mussel colonies, which established side-scanning as a viable tool for fine-scale identification of bottom features in shallow aquatic environments \cite{ref-greene2018,ref-powers2015}. Taken together, these studies suggest that a combination of DownScan and SideScan sonar mapping could support a multi-stage inspection workflow (DownScan and SideScan are Lowrance's names for downward-facing and left/right-facing sonar beams). Thus, dronar, coined from the integration of consumer-grade drone and sonar technology, has significant potential for non-invasive inspection of shallow-water infrastructure. 

However, flying drones, which have typically been used to tether the sonar transducer \cite{ref-diaz2022,ref-ruffell2021}, face numerous challenges in water canal inspection. First, they require certified pilots and regulatory approvals, which complicates operations in urban areas and near controlled airspace. The Arizona Canal, one of the major arteries in the SRP system, passes near multiple commercial airports, making airspace approval particularly complex in this application. Second, the drone's flight trajectory differs from that of the tethered sonar, making precise sonar positioning challenging; this is particularly important in narrow-water canals where the transducer must be centered for best results. Lastly, the limited battery endurance of flying drones prevents continuous inspection of long segments, with frequent recharging or battery swaps interrupting workflow and reducing efficiency.

To address these challenges, an autonomous unmanned surface vehicle (USV) equipped with a sonar transducer was developed to traverse the length of the canals and collect sonar data for analysis. USVs have experienced increasing adoption as mobile platforms for aquatic data collection and environmental monitoring in recent years \cite{ref-yamasaki2017,ref-duran2025}, offering the mobility and flexibility required for continuous survey operations without the airspace restrictions of aerial systems. 

This paper presents the design of the integrated dronar system, including mechanical and control aspects. Furthermore, data are presented that verify the operation and the utility of the sonar survey technique for non-invasive inspection of concrete water canals. 

\section{Materials and Methods}

\subsection{Dronar System}

\subsubsection{Sonar transducer}

The sonar transducer in this project was an Active Imaging HD 3-in-1 transducer, paired with an HDS Pro~9 sonar hub (both by Lowrance Electronics, Tulsa, OK). This system integrates high-resolution survey modes with onboard GPS, enabling georeferenced sonar returns and improving post-processing and mapping accuracy. The system's low power draw and compact form factor further support integration onto lightweight platforms. The transducer operates at 800~kHz and provides DownScan coverage up to $ \pm 30^{\circ}$ and SideScan coverage of up to $\pm 58^{\circ}$, making it well suited for canals, which are approximately 15--20~m wide and 1--2~m deep. When simultaneous DownScan and SideScan are used, a single lengthwise centerline pass along the canal segment can provide sonar coverage across most of the canal cross section. The sonar system automatically logs data to an internal SD card, minimizing operator workload during inspections.

\subsubsection{Integrated unmanned surface vehicle and sonar}

\begin{figure}[H]
\includegraphics[width=\textwidth]{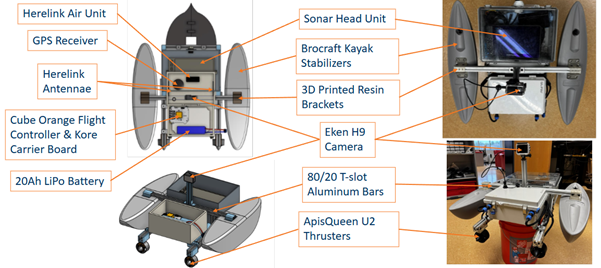}
\caption{Details of the custom-built unmanned surface vehicle and sonar, together referred to as dronar.\label{fig:usv}}
\end{figure}

The USV in this project was a custom-built twin-pontoon surface vehicle (Figure~\ref{fig:usv}). It consisted of two kayak stabilizer pontoons (Brocraft Kayak Outrigger) and an aluminum frame, and it was propelled using differential thrust from two electric thrusters (ApisQueen U2). The system was controlled using a CubePilot Cube Orange+ mounted on a Kore carrier board  (both by CubePilot Global Pty Ltd., Victoria, Australia). The Cube was programmed to run the rover firmware in ArduPilot, which is an open-source codebase for drone control \cite{ardupilot}. Remote control was maintained via a Herelink Air Unit (CubePilot Global Pty Ltd., Victoria, Australia), with extended range antennas that transmitted location coordinates (Here4 GPS module), speed, and live high-definition video (Eken H9 Camera). A 4-cell 14.8~V 20~Ah lithium polymer battery was used to power the USV. The entire vehicle weighed approximately 16~kg. Because the system is custom-built using modular commercial components and three-dimensional (3D)-printed brackets, it can be reconfigured for various payloads and deployment scenarios with minimal redesign.

The sonar transducer was mounted at the front of the USV using aluminum brackets secured to an aluminum crossbar. This mounting system suspended the transducer below the waterline and beneath the thrusters, reducing turbulence and improving survey image clarity. The USV also included two waterproof equipment boxes mounted on the deck to house the electronics. These latched boxes use rubber gasket seals and marine-grade fittings to ensure that internal electronics remain dry under operating conditions. The forward box contained the sonar head unit (Lowrance HDS Pro~9) and its dedicated external battery; the transducer cable exited through a sealed cable gland. The aft box contained the control and telemetry modules, as well as the main drone battery. This box featured waterproof glands for motor cables and sealed antenna pass-throughs for the Herelink extended-range system.

The dronar system was fully untethered, with no physical cables between the vehicle and the operator. Live sonar data was available on a tablet connected via WiFi, but this connection was range-limited and was not used during canal testing in this project. All sonar data and the logs of controller variables were stored locally for post-processing. The HD camera feed and telemetry were available in real-time and maintained strong signal quality throughout the 1.6 km test segment. In \emph{Manual} operation mode, the vehicle was remotely piloted at distances up to 1~km via the Herelink controller. If the connection was lost, the vehicle automatically transitioned to \emph{Loiter} mode and held position using GPS lock. In \emph{Auto} mode, the USV could be programmed to follow a predefined path independent of operator input, even after loss of radio signal.

\subsubsection{Dronar transportation and deployment}

\begin{figure}[H]
\centering
\includegraphics[height=4.7cm]{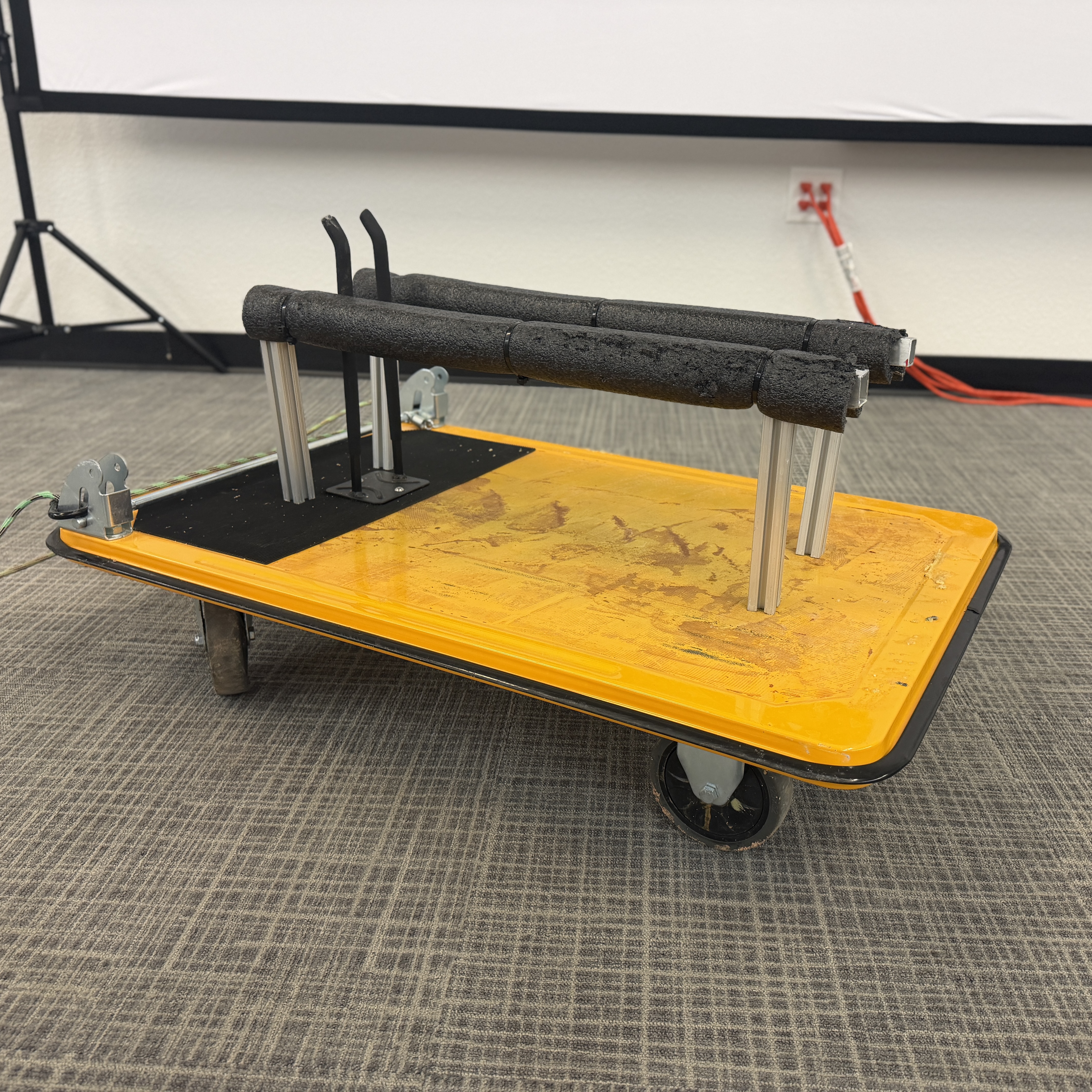}
\hspace{2pt}
\includegraphics[height=4.7cm]{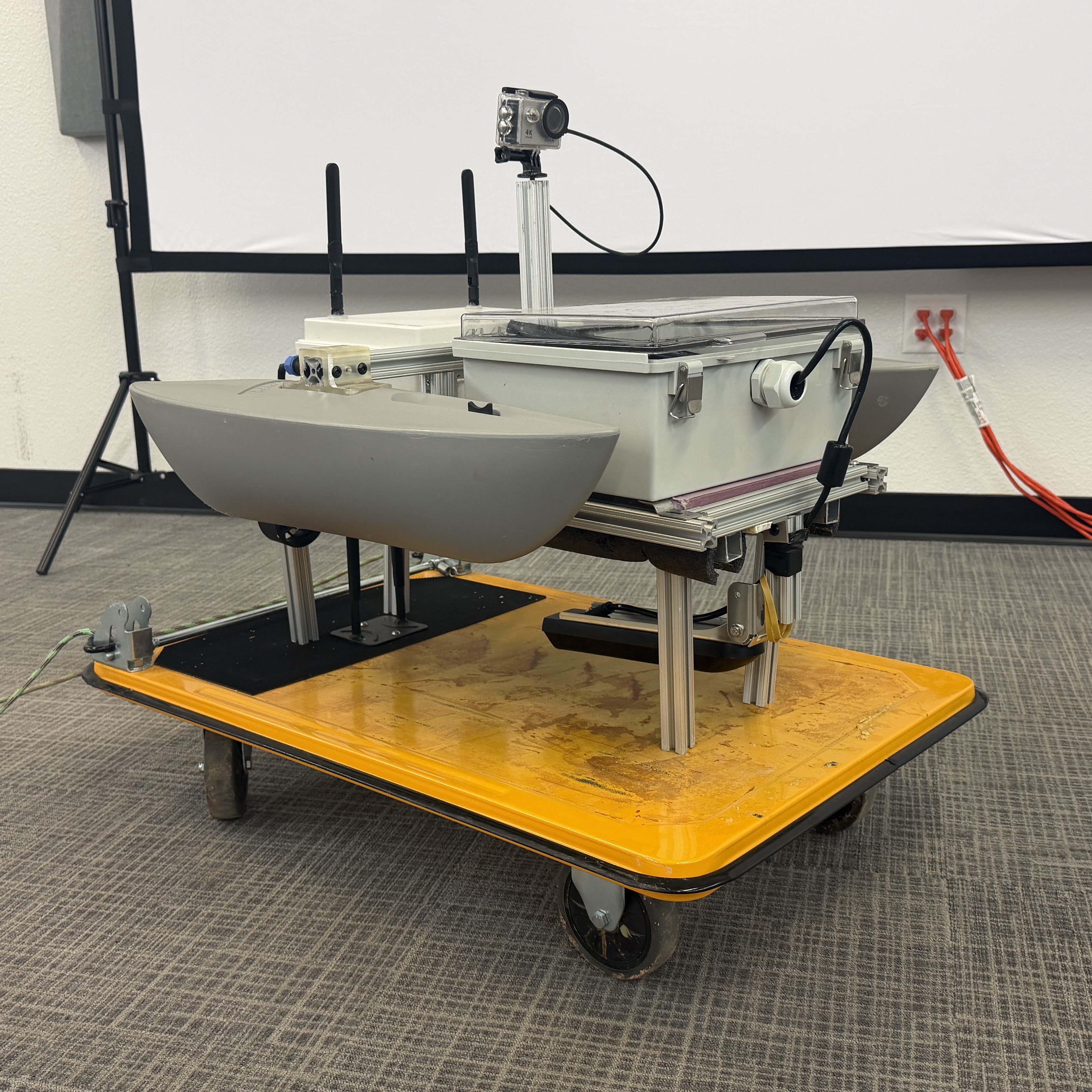}

\includegraphics[height=4.7cm]{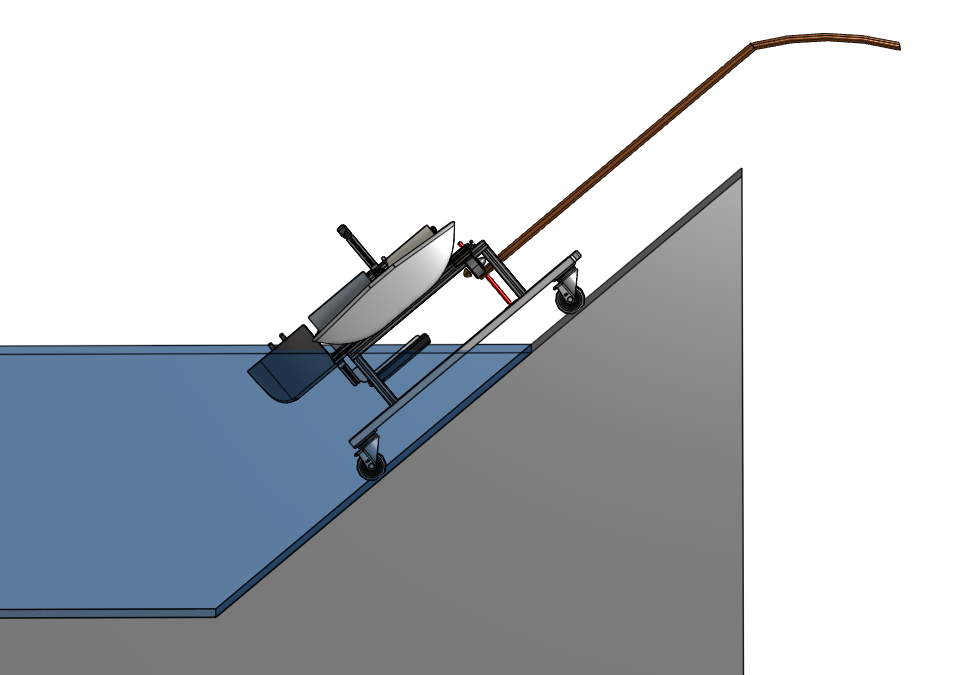}
\hspace{2pt}
\includegraphics[height=4.7cm]{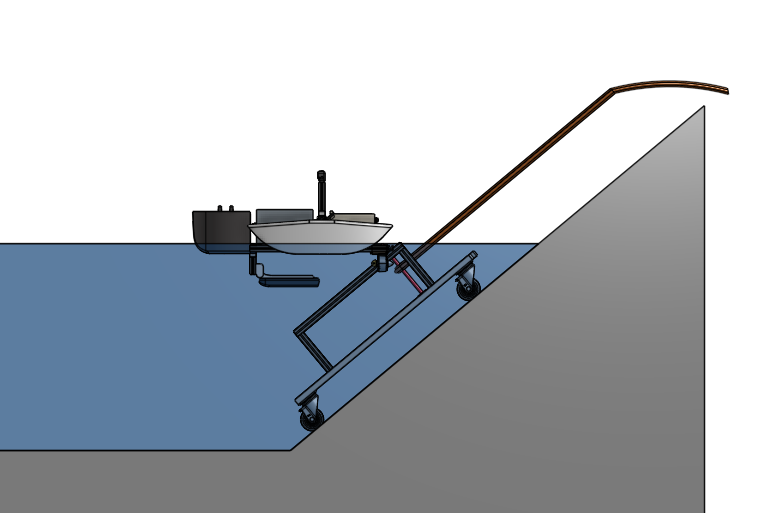}
\caption{(\textbf{a}) Dronar transport and deployment cart; (\textbf{b}) dronar system on deployment cart before launch; (\textbf{c}) CAD model of the USV being lowered on the cart; (\textbf{d}) CAD model of the USV being released from cart.\label{fig:cart}}
\end{figure}

To protect the low-hanging sonar transducer and simplify handling of the dronar during launch and recovery at the canal edge, a dual-purpose transport and deployment cart was developed with custom aluminum rails and a rear-mounted retention hook to cradle the dronar (Figure~\ref{fig:cart}). The cart doubled as a transport cradle, ensuring the transducer remained suspended in the air until water entry. The buoyancy of the cart was reduced by adding holes to the base and allowing water into the hollow cart platform. The cart facilitated rope-guided descent/ascent of the steep canal walls ($45^\circ$--$60^\circ$) during launch and recovery. Figure~\ref{fig:cart} shows the empty cart without the USV and the loaded cart holding the dronar prior to launch.

\subsection{Field Testing}

\subsubsection{Test 1: dronar operation verification}
\label{sec:firsttest}

\begin{figure}[H]
\includegraphics[width=10.0cm]{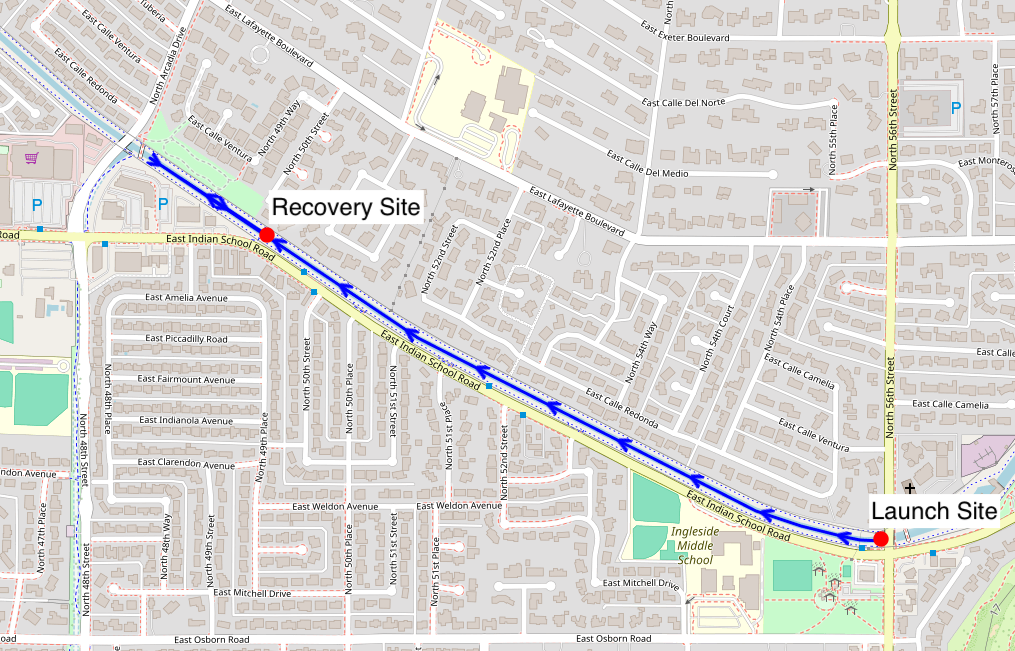}
\caption{Dronar proof-of-concept test location between N~56th~St and N~Arcadia~Dr in Phoenix, AZ.\label{fig:testmap}}
\end{figure}

A proof-of-concept canal survey was conducted on May~22, 2025, from approximately 4:30--6:00~pm on a 1.6~km segment of the Arizona Canal in Phoenix, AZ, between N~56th~St and N~Arcadia~Dr (Figure~\ref{fig:testmap}). The water flows westward, from 56th toward Arcadia. This segment of the canal was recently cleaned and inspected for canal liner damage during a canal dry-up, so there was limited to no sediment buildup or canal wall/bed deformation. The purpose of this test was to assess vehicle performance in moving water and to evaluate the full-scale use of the cart-based launch and recovery method.

During operation in the canal, the USV traveled the full 1.6~km segment in \emph{Manual} mode, heading downstream. After repositioning, the vehicle was set to \emph{Auto} mode and self-piloted upstream for a short distance along the same path. While navigating the canal, the Lowrance HDS Pro~9 sonar system stored data on an onboard SD card, which was post-processed using the sonarlight Python package \cite{sonarlight}. The USV logged location data throughout the mission.

For dronar deployment, the USV was secured on the transport and deployment cart, which was manually lowered with a rope into the canal down a sloped concrete boat access ramp at N~56th~Street (the launch site in Figure~\ref{fig:testmap}). Once the cart was submerged, canal flow pulled the vehicle free of the cart. During recovery along the canal wall after testing (recovery site in Figure~\ref{fig:testmap}), the USV was manually maneuvered into position upstream of the cart and perpendicular to the wall. The recovery wall had a slope of approximately 60$^{\circ}$. The USV was then allowed to drift downstream onto the cart, where it was pulled out once contact was confirmed.

\subsubsection{Test 2: pre- and post-maintenance surveys of a sedimented canal}

\begin{figure}[H]
\includegraphics[width=10.0cm]{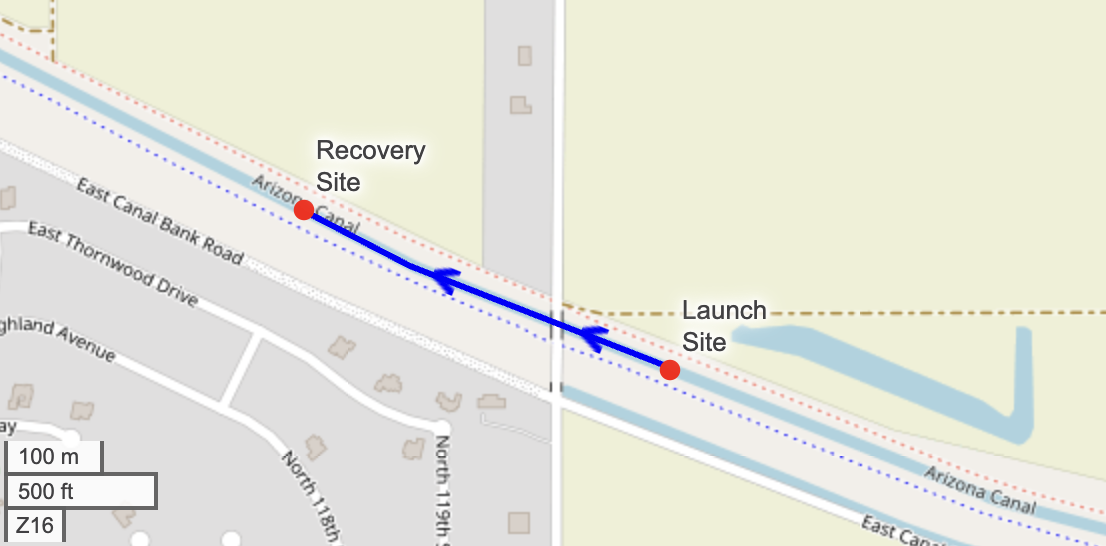}
\caption{Dronar pre- and post-maintenance test location at North Mesa Drive in Mesa, AZ.\label{fig:test2map}}
\end{figure}

A second set of tests was conducted to verify the repeatability and usability of sonar data to inspect canals with non-invasive techniques. For these tests, the segment of the Arizona Canal in Mesa, AZ, to the west of North Mesa Drive (Figure~\ref{fig:test2map}) was selected. This segment of the canal experienced heavy sedimentation after storms and was scheduled for maintenance. Therefore, repeated testing provided clear pre- and post-maintenance data to verify the sonar-based inspection technique.

The pre-maintenance survey was conducted on November~7, 2025, at approximately 10:00--11:00~am. The USV was controlled in \emph{Manual} mode, moving downstream, while the sonar and location data were recorded on the internal SD card for post-processing. The same launch and recovery procedure was used, with similar canal wall slope characteristics, with the addition of a gaff pole to help guide the USV onto the cart during recovery.

Soon after the first survey, this segment of the canal underwent thorough maintenance (cleaning and repair of canal liner damage). A post-maintenance survey was conducted on February~24, 2026, from approximately 10:00~am--12:00~pm, surveying the same segment (Figure~\ref{fig:test2map}). Six downstream-only runs were performed on the same test segment, two each of the following: \emph{Manual} mode at full forward speed, \emph{Manual} mode at canal water current speed (only side-to-side corrections by the operator), and \emph{Auto} mode. These data further enabled comparisons within the same run conditions for repeatability analysis. 

\section{Results}

\subsection{Operation of the Dronar}

\subsubsection{Speed, endurance, and test duration}

\begin{figure}[H]
\includegraphics[width=10.0cm]{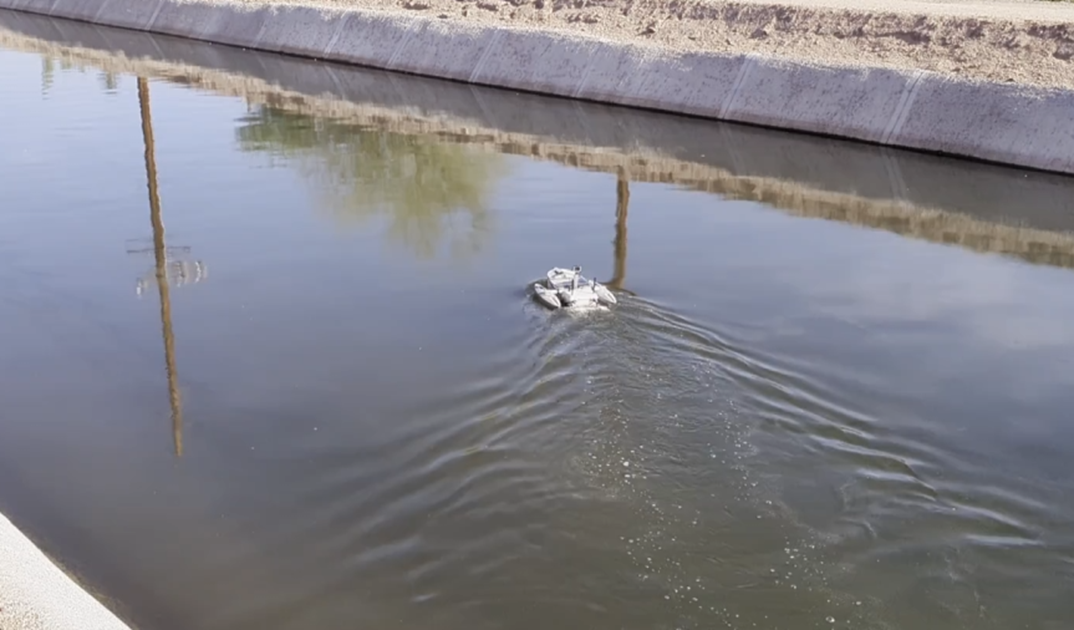}
\caption{Dronar USV traveling upstream in \textit{Manual} mode.\label{fig:upstream}}
\end{figure}

\textbf{Test 1: proof-of-concept (May~22, 2025).} The USV traveled the full 1.6~km downstream segment in \emph{Manual} mode, completing the run in approximately 25~minutes at 8~km/hr at full throttle. USV speed when traveling upstream was 1.6~km/hr at full throttle. Figure~\ref{fig:upstream} shows the USV traveling upstream at the end of this test. Battery consumption for the 1.6~km downstream segment was 38\%, with 62\% remaining upon completion. \textit{Loiter} mode was tested mid-run, which maintained dronar position within 1~m of the assigned GPS coordinate. During the test, there was no loss of telemetry.

\textbf{Test 2a: pre-maintenance (Nov.~7, 2025).} The USV was controlled in \emph{Manual} mode, moving downstream, while sonar and GPS data were logged on the onboard SD card for post-processing. Only one run of the approximately 400~m test segment was performed. The single run took approximately 10~minutes.

\textbf{Test 2b: post-maintenance (Feb.~24, 2026).} Six downstream-only runs were performed over the same test segment. Runs~1 and~2 were conducted with the dronar in \emph{Manual} mode at full throttle and took approximately 8 minutes each. Runs~3 and~4 were conducted with the dronar in \emph{Manual} mode at canal water current speed (side-to-side corrective inputs only) and took approximately 14~minutes each. Runs~1 through~4 were powered by a 16~Ah battery, which ended the four runs with 71\% charge remaining. Runs~5 and~6 were conducted with the dronar in \emph{Auto} mode and took approximately 7 minutes each, powered by a 20~Ah battery that ended with 75\% charge remaining. There was no loss of telemetry in any of the tests.

\subsubsection{Deployment and recovery}

\begin{figure}[H]
\centering
\includegraphics[height=7.0cm]{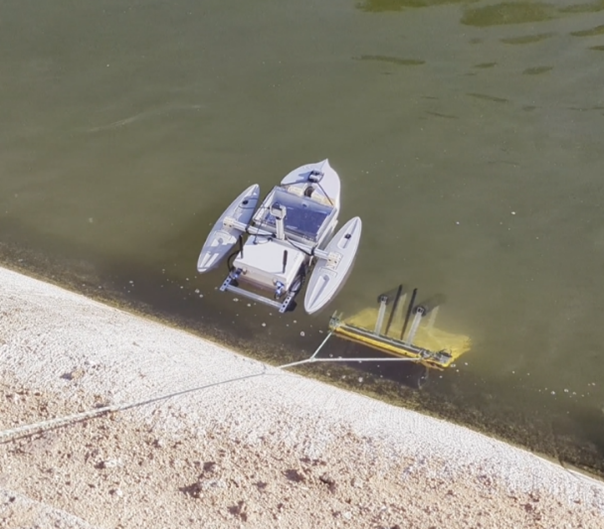}
\hspace{2pt}
\includegraphics[height=7.0cm]{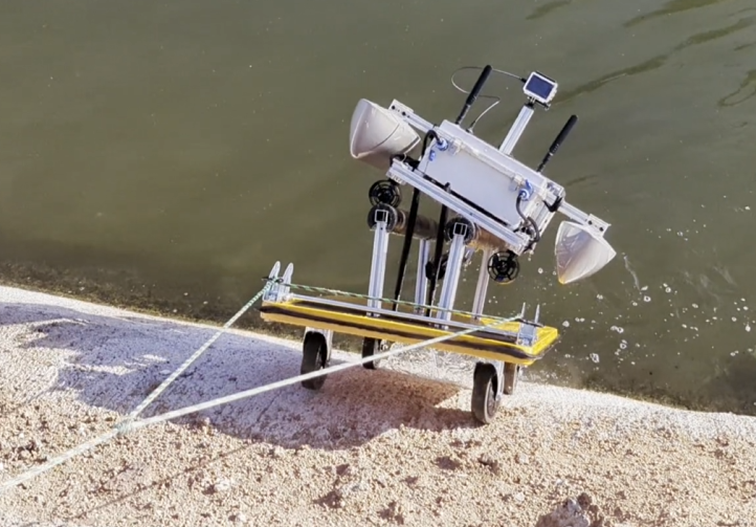}
\caption{(\textbf{a}) USV floating next to the cart prior to recovery; (\textbf{b}) USV recovery, with the left motor sitting on the rails but the cart hook still grabbing the frame and holding the USV secure.\label{fig:recovery}}
\end{figure}

In Test 1, the recovery method required four attempts to be successful due to cart drift and timing complexity, but the USV was eventually secured without damage. Figure~\ref{fig:recovery} shows the staged recovery operations. 

\begin{figure}[H]
\includegraphics[height=7.0cm]{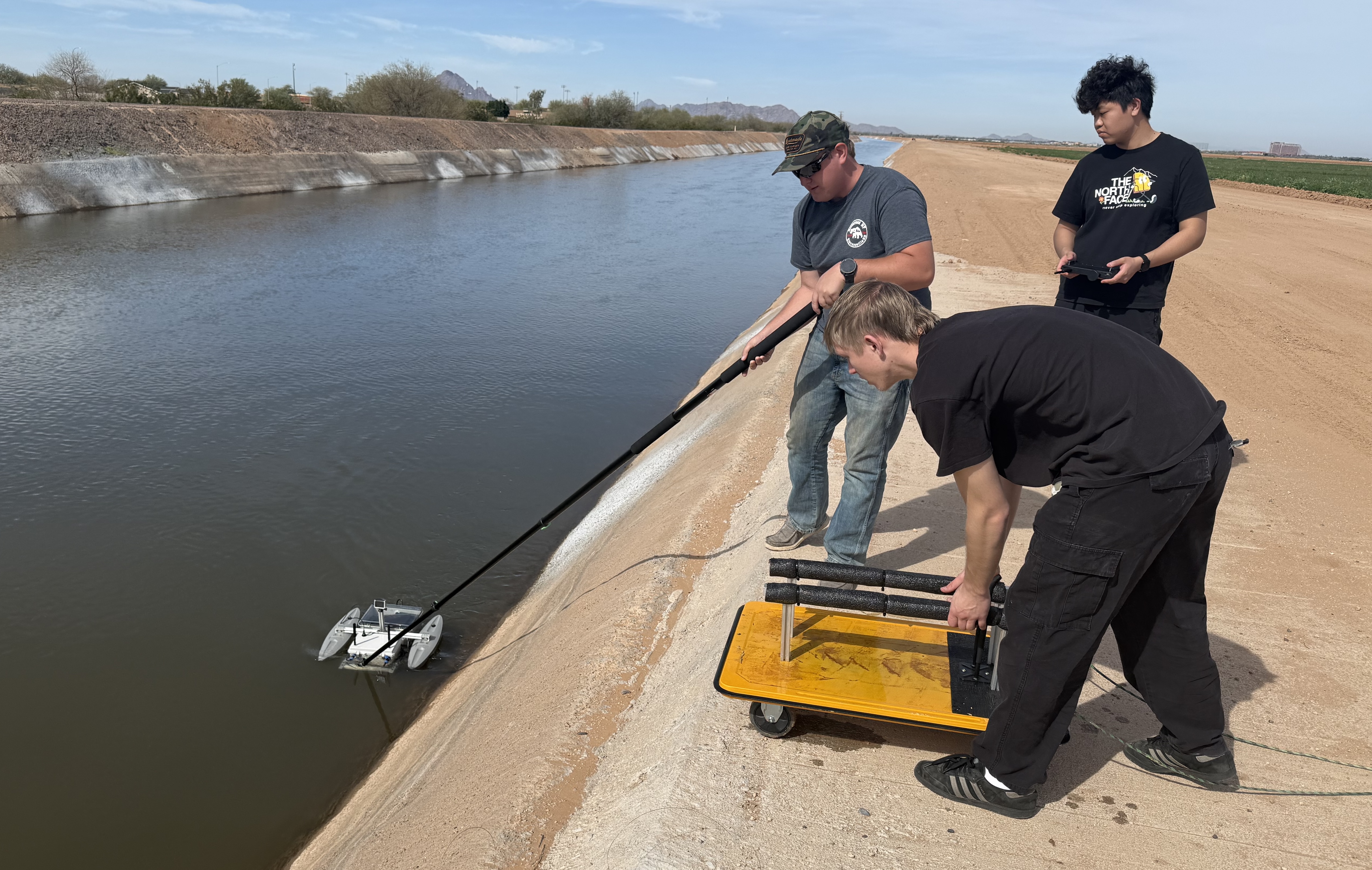}
\caption{USV being recovered utilizing the gaff pole.\label{fig:gaff}}
\end{figure}

The same recovery procedure was used for Test 2 with similar canal wall slope characteristics as Test 1. No boat ramp existed at this test site like at Test 1; the USV was launched and recovered from the same steep slope angle (Figure~\ref{fig:gaff}). During the post-maintenance tests, six sequential downstream runs were performed with recovery and redeployment occurring between each run. The first recovery took approximately 4 minutes, but each subsequent recovery was completed faster as the team gained experience with the procedure with recovery times reaching approximately a minute by the final run. 

\subsection{Sonar Data}

\begin{figure}[H]
\centering
\includegraphics[width=\textwidth]{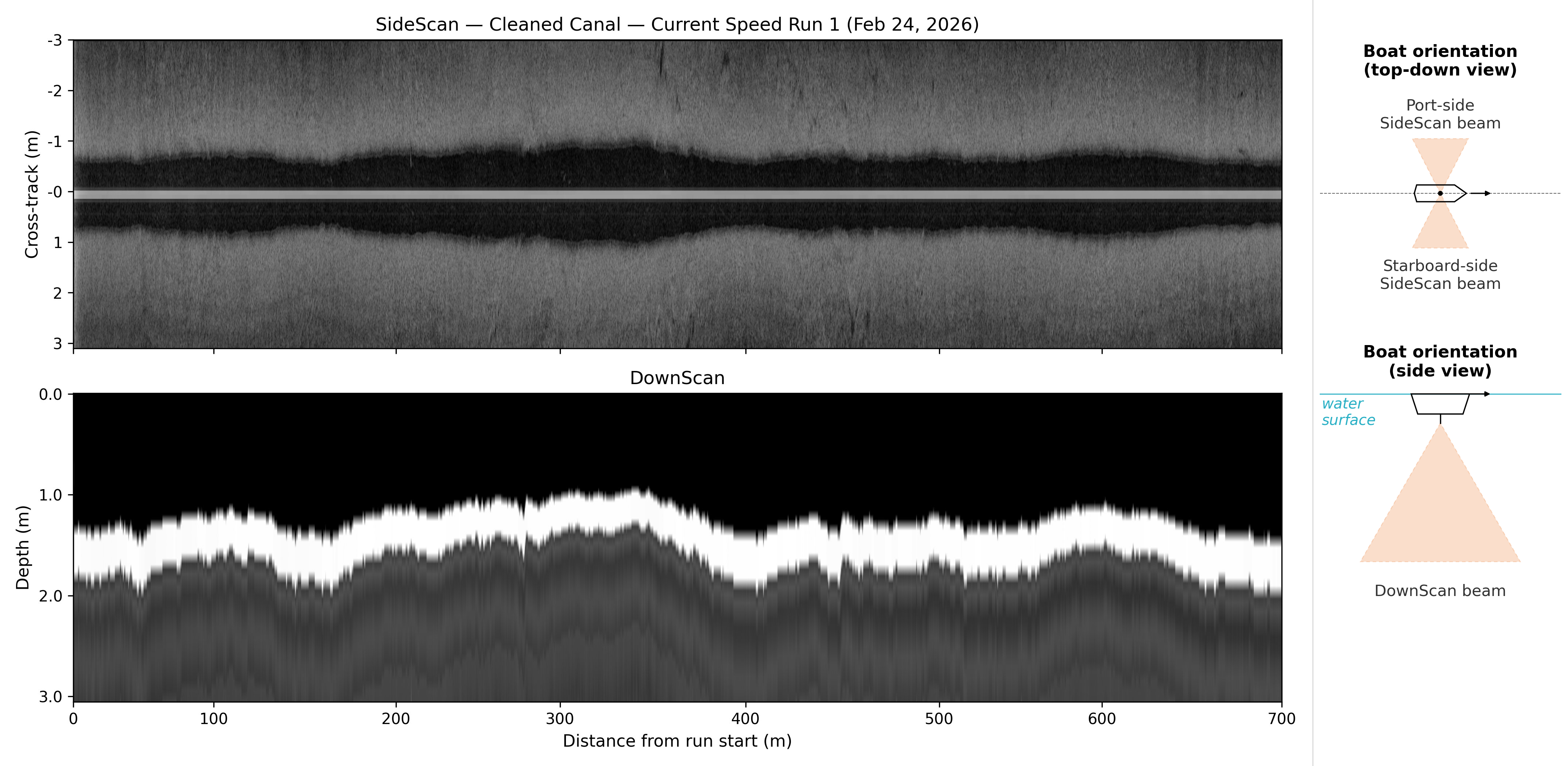}
\caption{SideScan (top) and DownScan (bottom) sonar imagery from the clean canal segment in Test 2b post-maintenance (Feb. 24, 2026). Both images are along the canal length; SideScan is a top-down view while DownScan is as if the user is looking from the side of the canal. Only DownScan data were used for the depth profile analyses in this study.\label{fig:sonar}}
\end{figure}

Figure~\ref{fig:sonar} shows an example of the detailed sonar data, imported by the sonarlight package and post-processed. The top image is the SideScan, and the bottom image is the DownScan. The images are uniform across the canal bed in this post-maintained segment. The depth profile analyses presented in the remainder of this section were derived exclusively from DownScan data; SideScan imagery is shown here for illustration of the data collected but was not analyzed quantitatively in this study.

\begin{figure}[H]
\centering
\includegraphics[width=\textwidth]{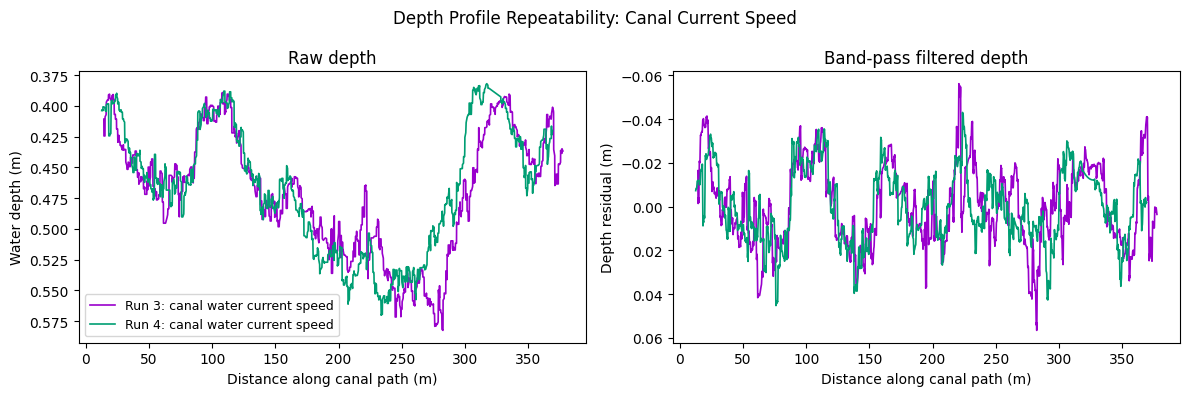}
\caption{DownScan depth profiles (m) for the two canal water current speed runs during Test 2b: post-maintenance (Feb. 24, 2026), showing raw and band-pass-filtered data.\label{fig:currentDepth}}
\end{figure}

To verify repeatability, Figure~\ref{fig:currentDepth} compares DownScan-derived depth data between different runs of the same canal segment in the post-maintenance survey in Test~2 (two runs conducted at the canal current speed). Both the raw and bandpass-filtered depth profiles of the two runs overlap closely throughout the segment. Comparisons between the other pairs of runs produced similarly well-matched depth profiles. Across all clean runs, repeated profiles were tracked within a maximum of 7~cm of each other, with the total depth range across the canal bed profile spanning approximately 12~cm. Some of the differences observed (e.g., the ``shift'' observed between 200--300~m in Figure~\ref{fig:currentDepth}) are due to inaccuracy in the calculation of the distance along the canal, rather than sonar data inaccuracy.

\begin{figure}[H]
\centering
\includegraphics[width=\textwidth]{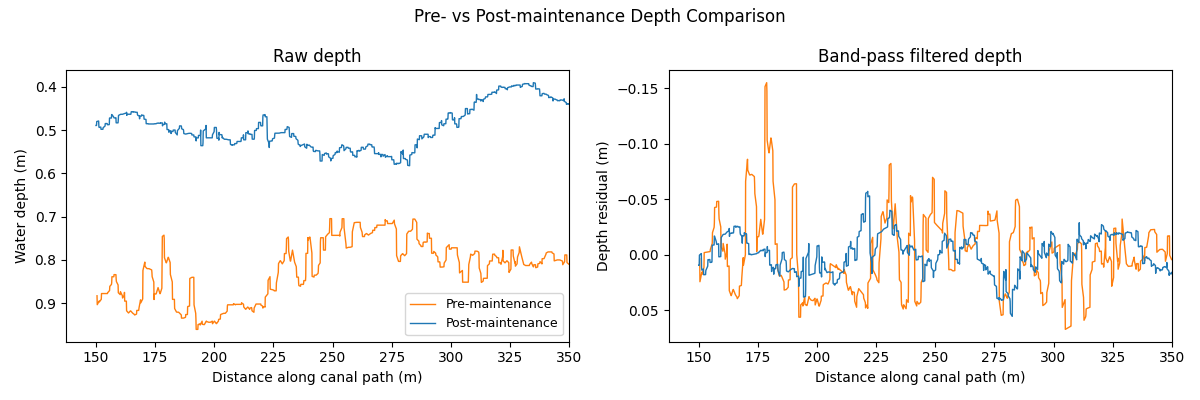}
\caption{DownScan depth profiles (m) along the Test~2 segment, shown as raw and bandpass filtered. Distance along canal path (m) is measured downstream from the Mesa Drive canal crossing; the x-axis is cropped to a representative subsection to show the contrast between sedimented and clean canal bed conditions.\label{fig:dirtyDepth}}
\end{figure}

With measurement repeatability confirmed, Figure~\ref{fig:dirtyDepth} compares the DownScan depth profiles for Test~2a: pre-maintenance (November 7, 2025) and Test~2b: post-maintenance (February 24, 2026) surveys. The raw water depth profiles (defined as the first acoustic return of the sonar, left panel of Figure~\ref{fig:dirtyDepth}) from the pre-maintenance survey were approximately 15~cm shallower than the profiles from the post-maintenance run over the same segment. The raw depth profiles contain both local bed-scale topography and a slowly varying background depth trend arising from canal grade, minor water surface slope, and small fluctuations in transducer depth throughout each run. A band-pass filter was applied to each depth profile to isolate intermediate-scale bed variations: spatial variations at wavelengths longer than 50~m were suppressed to remove the background depth trend, while variations shorter than 0.05~m were suppressed to remove high-frequency measurement noise. The filtered profiles express each measurement as a deviation from the local mean depth, placing the sedimented and cleaned segments on a common reference and making differences in canal bed condition directly visible. The filtered profiles (Figure~\ref{fig:dirtyDepth}, right panel) show the pre-maintenance trace exhibiting pronounced negative residuals at multiple locations along the segment, reaching as large as 15~cm, while the post-maintenance trace remains within approximately 5~cm or less of zero throughout. These negative excursions correspond to locations where sediment accumulation significantly reduced the sonar-measured depth below the local mean, confirming that the dronar detected spatially concentrated sediment deposits.

\section{Discussion}

\subsection{Operational Observations}

\subsubsection{Recovery procedure}

Recovery reliability improved substantially across Test 2, driven by both operator familiarity and the addition of a gaff pole. This suggests that a modest crew training protocol and standardized equipment list would be sufficient to achieve consistent recovery times in routine operational use.

\subsubsection{Thermal management}

No overheating or power delivery issues were observed in the propulsion system in any of the tests. During Test 1, which occurred in late afternoon in May in Phoenix's hot weather (air temperature around $35^{\circ}$C), the Herelink Air Unit was hot to the touch but did not cause any operation issues or produce any warnings. The Herelink Controller experienced overheating and issued a warning during the first canal test. While this issue did not cause a mission failure, sustained operations in high ambient temperatures may require thermal mitigation measures such as shielding, heat sinks, or scheduled cool-down intervals between runs. Test 2 was conducted in significantly cooler conditions (mornings in November and February). During these tests, there was no overheating from the controller.

\subsubsection{Speed and endurance}

Upstream travel speed was severely limited in Test 1 (1.6~km/hr upstream vs.\ 8~km/hr downstream). With no viable path for upstream survey under these canal flow conditions, Test 2 adopted a downstream-only approach, which proved effective for data collection and battery conservation.

The endurance data across all tests indicated that the dronar system had sufficient battery capacity for multi-kilometer downstream surveys. \textit{Auto} mode appears more power-intensive per run than \textit{Manual} mode, consuming a disproportionate share of battery capacity relative to run duration compared to the \textit{Manual} runs in Test 2b. Scaling to SRP's full 211~km canal system will require a battery management strategy, whether through multiple batteries, recharging stations, or segmented survey planning.

\subsection{Sonar Performance and Canal Condition Assessment}

The sonar system demonstrated its effectiveness for shallow-water sonar surveys across all canal tests. The sonar imagery from Test 1 showed no notable depth variations, consistent with the known clean condition of that canal segment, confirming that the system was operating as expected prior to the sediment detection tests. Comparing the pre- and post-maintenance DownScan data from Test 2 highlighted the ability of this technique to identify differences in canal bed conditions. Repeated surveys using the same throttle/mode configuration showed consistency between runs, establishing repeatability of the DownScan technique. These results confirmed that the dronar system was capable of identifying sediment buildup in water canals using DownScan depth profiling, and these results could be used to prioritize segments of the canal that require maintenance.  

\subsection{Limitations and Future Work}

Three limitations exist that should be addressed before operational deployment of the dronar system.

First, the current sonar validation was based on a comparison of two known canal conditions (sedimented vs. clean), and the dronar system can only identify the approximate size and depth of sediment deposits. Additional validation against other types of canal defects, including canal lining deformation, cracked concrete, sediment types, and foreign objects, is needed to fully characterize the system's detection capabilities.

Second, sonar data post-processing was manually completed using Python-based analysis scripts. Future work should focus on developing automated or semi-automated processing pipelines for rapid identification of features of interest, potentially including machine learning classification of canal bed conditions.

Third, this study's analyses relied only on DownScan data. Although the transducer's 60$^\circ$ beam width provides some lateral coverage, the shallow canal depths limit the sonar footprint, leaving canal margins and wall-adjacent areas unreachable by DownScan. SideScan data were collected during all three canal tests but were not analyzed in this work. Future studies should integrate SideScan analysis or adopt multi-pass survey patterns to provide lateral coverage of the canal cross section and enable feature characterization beyond nadir depth profiling.

\section{Conclusions}

This study investigated the feasibility of the dronar system, an off-the-shelf unmanned surface vehicle integrating consumer drone and sonar technology, for non-invasive canal bed inspection. Across multiple field tests on the Arizona Canal, the system was reliably deployed and recovered using a custom cart-based procedure, and downstream-only operation proved practical for multi-kilometer surveys with modest battery consumption. Depth profiles from a sedimented canal segment were consistently shallower than depth profiles from the same segment after cleaning, with offsets of up to 15~cm. Furthermore, repeated runs along the clean segment produced closely overlapping depth profiles, confirming measurement repeatability. These results demonstrated that the dronar can detect sediment accumulation and produce repeatable measurements under operational canal conditions. The dronar's use of affordable, modular hardware and ability to operate in active canals (without draining the water for inspection) positions it as a scalable system for use during routine inspection across a large concrete water canal network.


\section{Data Availability Statement}

The original contributions presented in this study are included in the article. Further inquiries can be directed to the corresponding author.

\section{Acknowledgments}

The authors would like to acknowledge the Salt River Project, specifically Andy Johnson and Ying Xu, for their continued funding and support behind the project. This research was funded by the NAU-SRP Cooperative Agreement, sponsored by Salt River Project, as well as a grant from the Water Infrastructure Finance Authority of Arizona.

\bibliography{references}

\end{document}